%% file: arxiv_version.tex
\documentclass[final]{cvpr}

\usepackage{listings}
\usepackage{times}
\usepackage{epsfig}
\usepackage{graphicx}
\usepackage{amsmath}
\usepackage{amssymb}
\usepackage{booktabs}
\usepackage{xcolor}
\usepackage{caption}
\usepackage{subcaption}
\input{math_commands.tex}

\usepackage{soul}
\usepackage{multirow}

\usepackage{listings}
\usepackage{xcolor}

\definecolor{codegreen}{rgb}{0,0.6,0}
\definecolor{codegray}{rgb}{0.5,0.5,0.5}
\definecolor{codepurple}{rgb}{0.58,0,0.82}
\definecolor{backcolour}{rgb}{1.0,1.0,1.0}

\lstdefinestyle{mystyle}{
    backgroundcolor=\color{backcolour},   
    commentstyle=\color{codegreen},
    keywordstyle=\color{magenta},
    numberstyle=\tiny\color{codegray},
    stringstyle=\color{codepurple},
    basicstyle=\ttfamily\footnotesize,
    breakatwhitespace=false,         
    breaklines=true,                 
    captionpos=b,                    
    keepspaces=true,                 
    numbers=left,                    
    numbersep=5pt,                  
    showspaces=false,                
    showstringspaces=false,
    showtabs=false,                  
    tabsize=2
}

\usepackage[pagebackref=true,breaklinks=true,colorlinks,bookmarks=false]{hyperref}

\def\cvprPaperID{8886} 
\def\confYear{CVPR 2021}

\newcommand{\birds}{CUB-200}

\definecolor{alexey}{rgb}{0.8, 0.0, 0.8}

\usepackage{cleveref}
\begin{document}

\title{Differentiable Patch Selection for Image Recognition}

\author{Jean-Baptiste Cordonnier$^{1\dagger}$\thanks{Work done during internship at Google Research.$^\dagger$ Equal contribution.}\;\quad Aravindh Mahendran$^{2\dagger}$\;\quad Alexey Dosovitskiy$^2$
\\
Dirk Weissenborn$^2$ \;\quad Jakob Uszkoreit$^2$ \;\quad Thomas Unterthiner$^2$
\\
$^1$EPFL, Switzerland \;\quad $^2$Google Research, Brain Team
\\
{\tt\small jean-baptiste.cordonnier@epfl.ch}\\
{\tt\small\{aravindhm,adosovitskiy,diwe,usz,unterthiner\}@google.com}
}

\maketitle

\begin{abstract}
Neural Networks require large amounts of memory and compute to process high resolution images, even when only a small part of the image is actually informative for the task at hand.
We propose a method based on a differentiable Top-K operator to select the most relevant parts of the input to efficiently process high resolution images.
Our method may be interfaced with any downstream neural network, is able to aggregate information from different patches in a flexible way, and allows the whole model to be trained end-to-end using backpropagation.
We show results for traffic sign recognition, inter-patch relationship reasoning, and fine-grained recognition without using object/part bounding box annotations during training.
\end{abstract}

\section{Introduction}
High-resolution imagery has become ubiquitous nowadays: both consumer devices and specialized sensors routinely capture images and videos with resolution in tens of megapixels.
Processing these high-quality images with computer vision models remains challenging: analyzing the images at full resolution can be prohibitively computationally expensive, while simply downsampling them before processing may remove important fine details and substantially hurt performance.
It would be desirable to save compute, while retaining the capability to recognize fine details.

\begin{figure}[t]
    \centering
    \includegraphics[width=\linewidth]{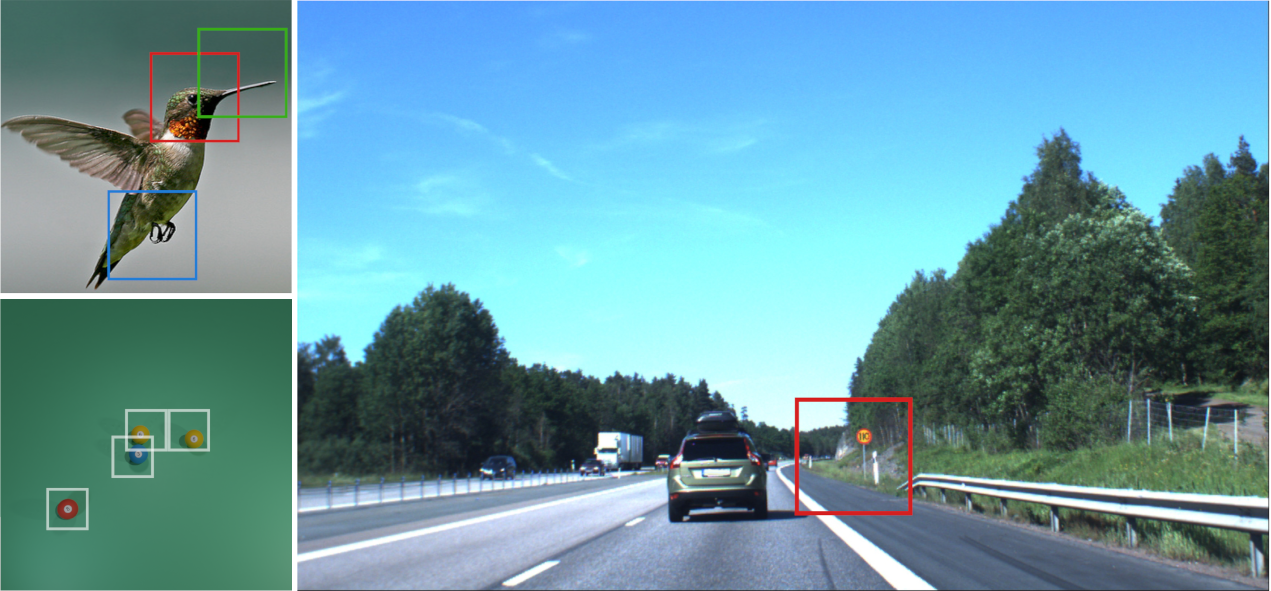}
    \caption{Examples of large images where patch extraction allows \emph{(top-left)} to focus on details for fine-grained recognition, \emph{(bottom-left)} to reason across patches, and \emph{(right)} to efficiently capture very localized information.}
    \label{fig:head_figure}
\end{figure}

Compute can be saved by exploiting the following property of many practical vision tasks: not all parts of the image are equally important for finding the answer.
Figure~\ref{fig:head_figure} shows examples of tasks where only a small fraction of the full image needs to be processed in detail.
Being able to quickly discard uninformative parts of the image would have several benefits.
It would reduce the overall computational and memory complexity of the model, and the regions of interest could be processed in more detail and by a more powerful model than otherwise.

Determining which parts of the image to retain and which to discard is usually nontrivial and highly task dependent.
In some applications the solution might be as simple as taking the center crop of the image, but in most cases relevant regions need to be detected first.
For instance, in a self-driving car setting, it would be permissible to ignore the sky, but all traffic signs in sight should be correctly identified and must not be ignored.
One may formulate this as follows: Given a regular grid of equally sized image patches, decide for each patch whether to process or discard it.
This decision is however discrete, which makes it unsuitable for end-to-end learning.

To overcome this limitation, inspired by the work of Katharopoulos \& Fleuret~\cite{attention_sampling}, we formulate patch selection as a ranking problem, where per-patch relevance scores are predicted by a small ConvNet and the Top $K$ scoring patches are selected for downstream processing. 
We make this end-to-end trainable with backpropagation using the perturbed maximum method of Berthet \textit{et. al}~\cite{berthet2020learning}. 
We present this as a generic module for patch selection. 
Our approach is most effective when the majority of patches in the image are irrelevant to the target, but the model a priory does not know where in the image the important patches are present. 
Hence, we do not aim to achieve image coverage such as in semantic segmentation and object detection.

In the remainder of this paper, we will formulate patch selection for image recognition as a Top-K selection problem in section~\ref{sec:patchextraction}, apply the perturbed maximum method to construct an end-to-end model trainable via backpropagation in section~\ref{sec:topkextraction}, and demonstrate wide applicability of this method via empirical results in three different domains: (1) street  sign recognition, (2) inter-patch relationship reasoning on synthetic data, and (3) fine-grained classification without using object/part bounding box annotations during training and evaluation (Section~\ref{sec:experiments}).

\section{Related Work}
\paragraph{Region proposal methods}: Several computer vision methods extract regions of interest from the image. Two stage object detection approaches, for instance, select regions of interest using region proposal networks~\cite{faster_rcnn} or hand crafted heuristics~\cite{Uijlings2013_selectivesearch,Girshick2015_fastrcnn}. Selected regions are later processed by a separate stage of the model. These methods use the non-differentiable RoI-Pooling~\cite{Girshick2015_fastrcnn} or the differentiable RoI-Align~\cite{He_2017_maskrcnn}. Such architectures require bounding box supervision to train large scale object detection models, whereas our experiments focus on a simpler setting and aim to train with weak supervision using only a single class label per image.

\paragraph{Soft attention}: In order to attend to specific parts of an image, an alternative approach is to occlude parts of the input by generating attention masks~\cite{show_attend_and_tell15}. While this helps the model focus on relevant features~\cite{sharma2015attention,Zheng_2017_ICCV, Wang_2017_CVPR,li2018tell}, become more interpretable~\cite{li2018tell}, or include external data such as image captions~\cite{Yang_2016_CVPR,Anderson_2018_CVPR}, the models will typically still process the whole image on a fixed input resolution. Thus they do not lead to any efficiency gains. Another approach~\cite{Chen_2016_CVPR} would be to process several image resolutions in parallel and use an attention mechanism to pick features from them. It is also possible to employ adhoc losses to extract meaningful patches~\cite{Ma2017_alamp}.

\paragraph{Multiple-Instance Learning}: A number of works use attention to solve Multiple-Instance Learning (MIL) problems, which are especially common in medical imaging, where images tend to be very large~\cite{ilse2018_attentionMIL,DBLP:conf/miccai/LiNLWCL19}. Here the goal is to label a set of related input samples, such as slices of organ scans or large images that are decomposed into patches. While, for example, the method of Ilse \textit{et. al}~\cite{ilse2018_attentionMIL} can be used to identify the most relevant patches, this is not leveraged to make computation more efficient, as all image patches are processed in equal detail by this method.

\paragraph{Sequential ``glimpses''}: There is a long line of work that sequentially processes a sequence of patches (``glimpses''), from a network, until they settle on the most relevant ones~\cite{Schmidhuber91attention,Fu2017_CVPR}. These methods often rely on Reinforcement Learning to train non-differentiable attention mechanisms~\cite{mnih2014_draw,ba-attention-2015, Li2017_ICCVW, elsayed2019saccader}, which typically makes them difficult to train.
The Spatial Transformers~\cite{spatial_transformer} on the other hand can be deployed as a differentiable attention mechanism, for example for fine-grained recognition of bird species. These have been applied sequentially to extract several regions of interest~\cite{eslami2016neurips,kuen2016recurrent} as part of recurrent neural networks. Training spatial transformers on large images can, however, be difficult because the gradients with respect to the transformation parameters are an accumulation of gradients of sub-pixel bilinear interpolation which can be very local. Angles \textit{et. al}~\cite{angles2021mist} overcome these limitations to train a multiple instance spatial transformers by lifting non-differentiable Top-K by introducing an auxiliary function that creates a heat-map given a set of interest points. Our method, on the other hand, avoids these limitations by computing gradients with respect to all patches in every backward step.

\paragraph{Attention Sampling}: Most related to our current work are differentiable methods that sample patches, specifically Attention Sampling (ATS)~\cite{attention_sampling}. ATS requires that the output of the network $f(x; \theta)$ be an expected value over embeddings of all possible patches $\tP$. That is, $f(x, \theta) = \mathbb{E}_{p \sim Z(\tP)}\left[f(p; \theta)\right]$. Using a Monte Carlo approximation of the expectation, $f$ is only applied on a small number of patches sampled according to the distribution $Z(\tP)$. That is, $f(x, \theta) = \frac{1}{K}\sum_{i=1}^K f\left(p_i; \theta\right)$. Thus ATS is restricted to a simplistic average pooling scheme for aggregating information from the extracted patch embeddings. Our method, on the other hand, solves the patch sampling problem using a differentiable Top-K operation, which allows us to combine the per-patch information in a flexible manner.

\paragraph{Differentiable Top-K}: The subset sampling operation can be implemented by expanding the Gumbel-Softmax trick~\cite{xie2019reparameterizable} or based on optimal-transport formulations for ranking and sorting~\cite{smooth_topk_xie, diff_rank_sort_cuturi}, the latter of which was recently made significantly faster by Blondel~\textit{et. al}~\cite{fast_diff_sorting_blondel}. Our work uses perturbed optimizers~\cite{berthet2020learning} to make a Top-K differentiable, which we found performed better than the Sinkhorn operator~\cite{smooth_topk_xie}.

\section{Model overview}\label{sec:patchextraction}

\begin{figure*}
    \centering
    \includegraphics[width=\linewidth]{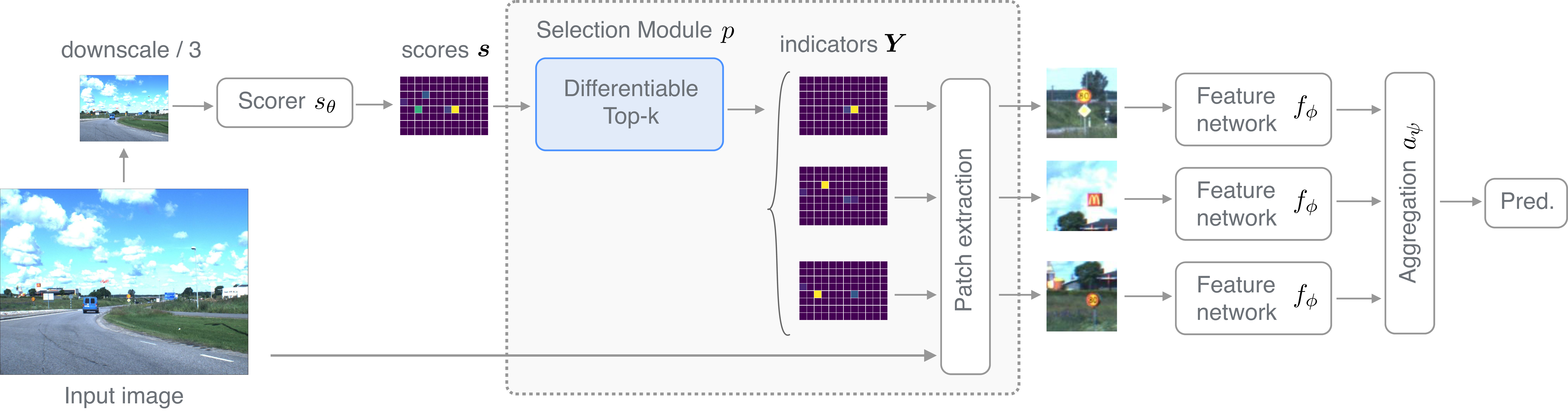}
    \caption{The differentiable Top-K layer uses the scores output by $s_\theta$ to extract patches that can be processed by an \emph{arbitrary} downstream network.
    The whole model is trained end-to-end without any adhoc loss to train the selection module.}
    \label{fig:birds_architecture}
\end{figure*}

Our model processes high resolution images by scoring, selecting then processing, some regions of interest. As illustrated in Figure~\Ref{fig:birds_architecture}, the model consists of a scorer network $s_\theta$, a patch selection module $p$, a feature network $f_\phi$ and an aggregation network $a_\psi$, where $\theta, \phi$, and $\psi$ are learnable parameters. Attention sampling (ATS)~\cite{attention_sampling} is a special case of this where aggregation is an average pool and patch selection is implemented using discrete sampling.
To describe the model in one sentence: The scorer scores patches, the patch selection module selects a subset of those, the feature network computes embeddings for each patch, and the aggregation network combines these embeddings to make a final prediction. These modules are described in more detail below.

The \textbf{scorer network} $s_\theta$ predicts a relevance score for each region of the image with respect to the task at hand. This network is meant to be shallow and typically operates on downscaled images for applications where the original image size is too big. It outputs relevance scores, $\mS = s_\theta(\tX) \in \R^{h \times w}$, which constitute a $h \times w$ grid. This can be easily generalized to a scorer network that predicts scores for patches at different scales and aspect ratios, much like the region proposal network in Faster-RCNN~\cite{faster_rcnn}. Crucially though, our model selects only a small number of regions of interest (e.g. 10), when compared to the 2000 used in Faster-RCNN as necessitated by the high recall requirement for object detection.
A \textbf{selection module} $p(\tX, \mS)$ uses this information to extract the $K$ most relevant patches from the image.
The output of the selection module are $K$ patches of dimension $P_h\times P_w$ denoted by $\tilde \tX \in \R^{K\times P_h \times P_w \times C}$\footnote{Patch sizes need not coincide with the scorer network's receptive field size. For example, the scorer network might score $32 \times 32$ pixel patches, while the selection module may extract patches of size $50 \times 50$ pixels.}.
To simplify the notation, we use a single patch size but in practice (see Section~\ref{ssec:birds}) one can also consider patches at different resolutions and aspect ratios.
For each individual patch, a \textbf{feature network} $f_\phi$ calculates a $D_h$-dimensional representation, which for all patches together is a matrix $\mH \in \R^{K\times D_h}$. $f_\phi$ is typically a computationally large and expensive network. One may use different feature networks for different patches but we did not explore this direction.
Finally, an \textbf{aggregation network} $a_\psi$ pools this information into the model output $\vy = a_\psi(\mH) \in \R^{D_o}$. This network can be a simple mean pooling operation as in ATS, or an equally simple max pooling operation, or a sophisticated transformer network or anything in between.

Several prior works can be cast in this manner: 
\emph{Spatial transformers}~\cite{spatial_transformer} have localisation networks that are similar to our scorer network, albeit instead of scoring patches they predict localization parameters such as an affine transformation. Thus scoring and patch selection are effectively done together. Their grid sampler can be interpreted as a patch extraction module.
\emph{Attention sampling}~\cite{attention_sampling} sample patches (with or without replacement) according to normalized scores output by a scorer network and use a ResNet for feature extraction $f_\phi$. The aggregation network $a_\psi$ is restricted to taking the average of the patch embeddings to be able to back propagate through the discrete sampling operation.
\emph{Vision transformer}~\cite{dosovitskiy2020image,Cordonnier2020On}: the patch selection module is exhaustive and extracts all $P\times P$ patches in the image with a stride $P$. The feature network $f_\phi$ and the aggregation network $a_\psi$ are merged into a large transformer that process all the flattened patches with self-attention and output the representation of the CLS token or an average of the tokens' representations.

\section{Patch selection as differentiable Top-K}\label{sec:patchselect}
We aim to train our model end-to-end without introducing any auxiliary losses for individual components. The core technical novelty of our method is the patch selection module.
Roughly speaking, we formalize patch selection as the Top-K problem: $\operatorname{Top-K}\left(x \in \mathbb{R}^N\right) = y \in \mathbb{N}^K$ where $y$ contains the indices of the $K$ largest entries in $x$. We then apply the perturbed maximum method~\cite{berthet2020learning} to construct our differentiable patch selection module.

\subsection{Patch selection as Top-K}
Given scores from the scorer network, we select the $K$ highest scores and extract the corresponding patches.
Specifically, given scores $\mS \in \R^{h\times w}$ for $N=h \times w$ patches, Top-K returns the indices of the $K$ most salient patches in the image. For reasons that will be clear in the next subsection, we define Top-K such that the indices are sorted. That is, $y_1 < y_2 < \cdots < y_K$. Without this constraint Top-K's output could be permuted arbitrarily breaking the perturbed optimizer method. Furthermore, we represent $y_i$'s as one-hot $N$ dimensional indicator vectors $\lbrace I_{y_1}, I_{y_2}, \cdots, I_{y_K} \rbrace$.
The intuition here is that if we had a large tensor of all patches in the image, we could ``extract'' all chosen patches using a single matrix tensor multiplication.
Specifically, given all patches, $\tP \in \R^{N \times P_h \times P_w \times C}$, and $\mY = \left[I_{y_1}, I_{y_2}, \cdots, I_{y_K}\right] \in \lbrace 0, 1\rbrace^{N \times K}$, patches may be extracted as $\tilde \tX = \mY^\top \tP$~\footnote{In practice, we use \texttt{jax.lax.scan} as this is memory efficient.}.
This operation is non-differentiable because both Top-K and one-hot operations are non-differentiable.

\subsection{Differentiable Top-K}\label{sec:topkextraction}
To learn the parameters of the scorer network using backpropagation, we need to differentiate through the patch selection method. We employ the perturbed maximum method~\cite{berthet2020learning} for this purpose. Given a non-differentiable module, whose forward pass can be represented as a linear program of the form
\begin{equation}
    \argmax_{\mY \in \mathcal{C}} \langle \mY, \bf{\eta} \rangle\,.
\end{equation}
with inputs $\eta$, optimization variable $\mY$, and convex polytope constraint set $\mathcal{C}$, the perturbed maximum method defines a differentiable module with forward and backward operations as below.
\paragraph{Forward:} Sample uniform Gaussian noise $Z$ and perturb the input $\eta$, to generate several perturbed inputs. Solve the linear program for each of these, and then average their results.
\begin{equation}
    \mY_{\sigma} = \mathbb{E_Z}\left[\argmax_{\mY \in \mathcal{C}} \langle \mY, \bf{\eta} + \sigma Z \rangle \right]
\end{equation}
where $\sigma$ is a hyper-parameter. In practice one computes the expectation using an empirical mean with $n$ independent samples for $Z$. We fix $n=500$ in all our experiments and tune $\sigma \in \{0.01, 0.05, 0.5\}$. This does not require one to solve $500$ linear programs in every forward pass, instead as the linear program is chosen to be equivalent to Top-K, we run the Top-K algorithm $500$ times, one for each perturbed input, which is very fast in practice.
\paragraph{Backward:} Following \cite{Abernethy2016perturbations}, the Jacobian associated with the above forward pass is
\begin{equation}
J_\vs \mY = \E_Z \left[\argmax_{\mY \in \mathcal{C}} \langle \mY, \bf{\eta} + \sigma Z \rangle Z^T / \sigma \right] .
\end{equation}
The equations above have been simplified for the special case of normal distributed $Z$\footnote{Other distributions could be used. Please see~\cite{berthet2020learning} for details}.

Patch selection as Top-K with sorted indices is equivalent to the following linear program. 
\begin{equation}
\max_{\mY \in \mathcal{C}} \langle \mY, \vs \vone^\top  \rangle\,.
\end{equation}
where $\vs$ are scores output by the scorer network. $\vs \vone^\top \in \mathbb{R}^{N \times K}$ are scores replicated $K$ times. $\langle \rangle$ flattens the matrices before taking a dot product. The constraint set is define as
\begin{align}
\label{eqn:constraitset}
    \mathcal{C} = \big\{\mY \in \R^{N\times K}: & ~\mY_{n,k} \geq 0,\vone^\top \mY = \vone, \mY \vone  \leq \vone,\\
    &\sum_{i\in[N]} i\mY_{i,k} < \sum_{j\in[N]} j\mY_{j,k'}~ \forall k < k'\big\}\,.\nonumber
\end{align}
Note how $\mY$ has the same shape as the concatenation of indicator vectors $\mY$ in the previous subsection. They serve the same purpose. The first conditions encourages that $\mY$ is an assignment where each of the columns has a total weight of one. The last condition results in sorting the indices. This linear program has infinitely many optimal solutions, one of which is the required integer solution corresponding to the index-sorted ``Top-K'' operator.

Note that in theory, noise should be applied to $\vs \vone^\top$. The equivalence between this linear program and Top-K crucially relies on all columns of $\vs \vone^\top + \sigma Z$ being identical. We therefore apply noise directly to $\vs$. Our experiments show that this departure from theory works in practice. We normalize the scorer output $\vs$ to lie in $[0, 1]$ with a small $\epsilon=10^{-5}$ to avoid any division by zero.

\begin{figure}[b]
    \centering
    \includegraphics[width=\linewidth]{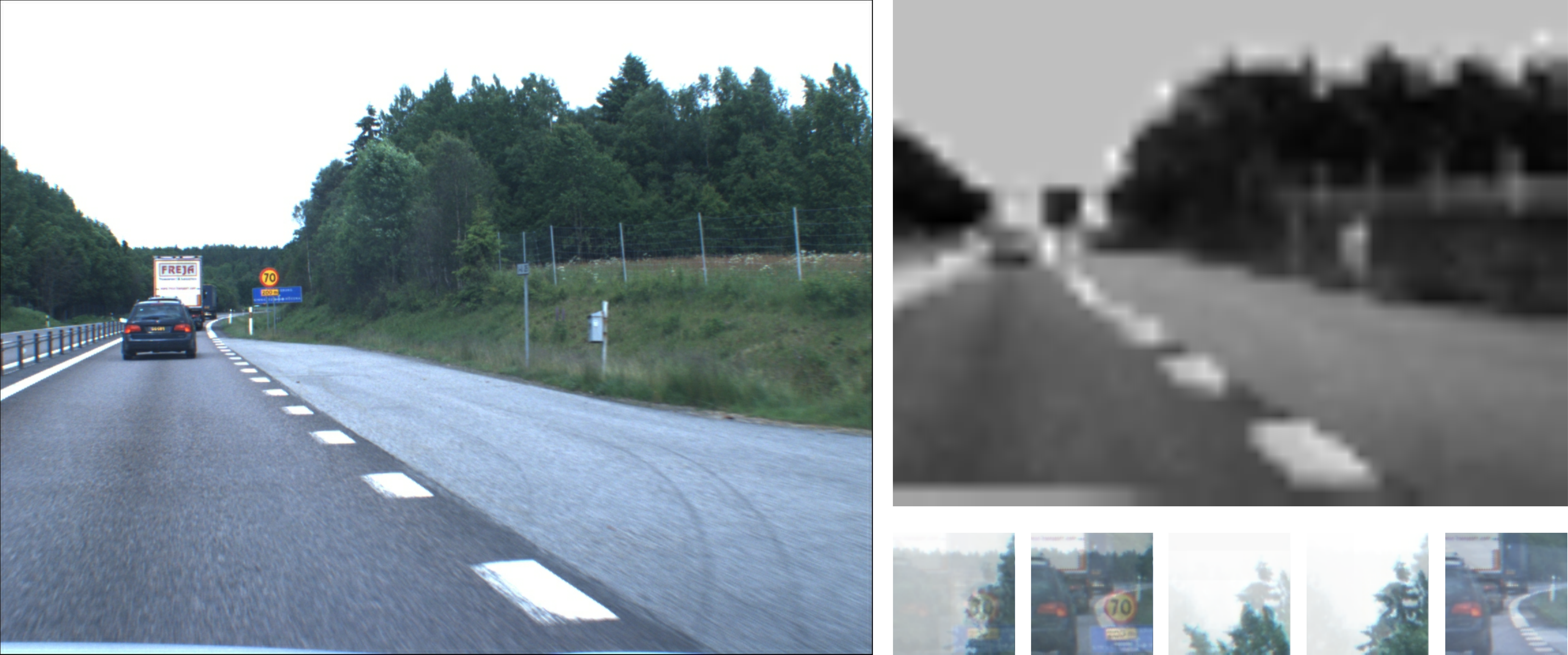}
    \caption{Scores computed at initialization (\emph{top-right}) and extracted patches (\emph{bottom-right}) displaying interpolation.}
    \label{fig:at_initialization}
\end{figure}

While Top-K for each perturbed input will result in one-hot indicators $\mY$, their perturbed average may be far from one-hot. Thus early in training, when the scores are still non-decisive, extracted patches resemble a weighted average over all image patches (Figure~\ref{fig:at_initialization}). Mixing images like this might contribute to model generalization~\cite{mixup}.
Another side benefit is that backpropagated gradients take into account all image patches and can update their weights from the very first step. A discrete sampler on the other hand is solving a combinatorial search over patches and gradients may be non-informative until the right patches have been sampled consistently for a few iterations of training.
The average can, however, become meaningless if the Top-K solver returned a different permutation of the $K$ most salient patches for each perturbed input. Sorting of indices overcomes this issue and is an integral part of our ``Top-K'' operator.

For efficiency reasons we use hard top-K during inference. Hard top-K processes 3-10\% more images per second because only a single Top-K operation has to be performed (instead of $n$ perturbed repetitions) and patch extraction by slicing the tensor is more efficient than weighted combination with the indicator vectors. Furthermore, hard Top-K is deterministic which might be a desirable property at inference. However, using hard top-K at inference results in a train-test gap. To bridge this we linearly decay $\sigma$ to zero during training. At $\sigma = 0$, no noise is added and the differentiable top-k operation is numerically identical to hard top-K. The gradients flowing into the scorer network vanish at $\sigma = 0$.

We also experimented with another differentiable Top-K formulation based on Sinkhorn operators~\cite{smooth_topk_xie}. We found the above perturbed-optimizer formulation to give superior results. Sinkhorn based experiments are therefore deferred to the appendix (Section D).

\section{Experiments and results}\label{sec:experiments}
Differentiable patch selection is a generic tool that can benefit a variety of computer vision problems. In this work, we focus on selecting a small number of patches from high resolution images for image classification.

\subsection{Traffic Signs Recognition}
\label{ssec:trafficsigns}

Our first task is to recognize speed limits signs in large images.
This is a key task to enable autonomous driving and is a natural fit for our method since the relevant pixels in the image are very localized and the model must rely on the high resolution images to read speed indications at significant distances.
We use the Swedish traffic signs dataset \cite{Larsson2011trafficsigns}, replicating the setup of \cite{attention_sampling} for their Attention Sampling (ATS) method.
ATS uses a subset of the dataset consisting of 747 training images and 684 test images of dimension $960 \times 1280$ pixels, and the goal is to classify whether each image contains a limit sign of 50, 70 or 80 kilometers per hour or no speed limit.
We apply the same data augmentation as \cite{attention_sampling}, specifically a random translation and a random affine color scaling per image.
For a fair comparison, we use ATS's scorer: the scorer is a 4-layers CNN followed by a stride 8 max-pooling layer. We apply the scorer on a $3\times$ downscaled version of the image. We obtain scores for $39\times 52 = 2028$ candidate patches of dimensions $100\times 100$ and select $K \in \{5, 10\}$ of them.
The feature network $f_\phi$ is the modified thin ResNet used by ATS. We use mean-pooling to aggregate per-patch representations.

\begin{figure}
    \centering
    \includegraphics[width=0.3\textwidth]{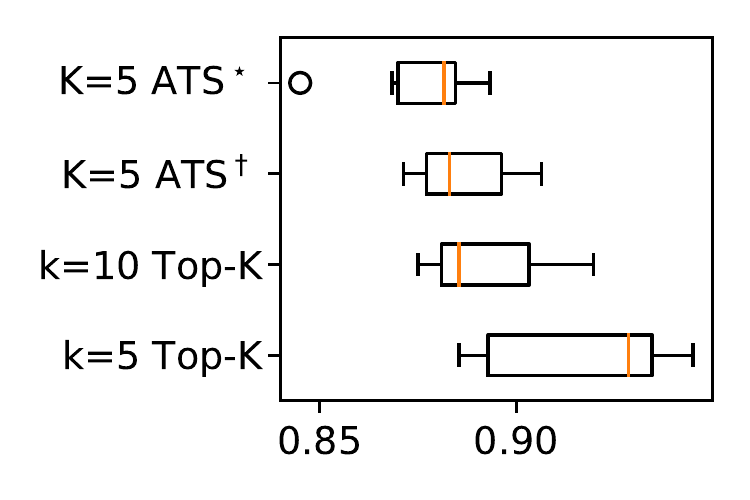}
    \caption{Performance on the traffic signs dataset across 9 repeats.}
    \label{fig:cvpr21:cr:trafficsigns_results}
\end{figure}

Results in \Cref{tab:trafficsigns_results} show that we match the mean performance reported by ATS in their paper~\cite{attention_sampling}, albeit with a larger variance.
We tried reproducing ATS results using their publicly released code, but were unsuccessful. The rows marked as $\mathrm{ATS}^\dagger$ and $\mathrm{ATS}^\star$ show the best stable results we could obtain. These closely follow the hyper-parameters reported in their paper using an entropy regularization of $0.05$ and training for $300k$ iterations. We compare the performance of our method and these reproduced ATS results in the box and whiskers plot of~\cref{fig:cvpr21:cr:trafficsigns_results}. ATS can achieve higher accuracy with shorter training schedules by using a lower entropy regularization coefficient of $0.01$. However, in this setting, approximately $20-25\%$ of runs fail on the test set with accuracy less than approximately 60\%.

We ablate against not having a patch selection procedure and directly apply the feature network CNN to the full image.
Beside wasting computation over constant parts of images (e.g. the blue sky), this small ResNet completely overfits the training set: reaching 100\% training accuracy while not achieving better accuracy than just predicting the majority class on test.
It seems that the inductive bias introduced by patch extraction is crucial in this very low data regime.

\begin{table}
\centering
\begin{tabular}{@{}llc@{}}
\toprule
                           & K & test acc. [\%]\\
\midrule
Top-K (Ours)               & 5  &$91.7 \pm 2.2$ \\
Top-K (Ours)               & 10  &$89.3 \pm 1.4$ \\
\midrule
ATS~\cite{attention_sampling}     & 5       & $91.1 \pm 0.2$\\
ATS~\cite{attention_sampling}     & 10      & $90.5 \pm 0.8$\\
$\mathrm{ATS}^\dagger$            & 5       & $88.6 \pm 1.1$\\
$\mathrm{ATS}^\star$              & 5       & $87.6 \pm 1.4$\\
\midrule
CNN                               & -       & $63.0 \pm 2.6$\\
\bottomrule
\end{tabular}
\caption{Performance on the traffic signs dataset. ATS results are reported from~\cite{attention_sampling}. We report the mean and the standard deviation of 9 runs except for CNN which uses 5 runs. $\mathrm{ATS}^\star$: Our reproduction of ATS following the hyper-parameters in~\cite{attention_sampling}, $\mathrm{ATS}^\dagger$: the same but with learning rate drop at 12000 epochs.}
\label{tab:trafficsigns_results}
\end{table}

We manually investigated the mistakes made by our model. Anecdotally the scoring network was able to extract the most relevant patches in most cases. Its only failure mode was a tendency to extract false-positive patches that exhibit the same colors as the traffic signs in question. Most miss-classifications are likely due to either the feature- or the aggregation-network. One could potentially further improve our results by pre-training the feature network.

\begin{figure}[h!]
    \centering
    \includegraphics[width=\linewidth]{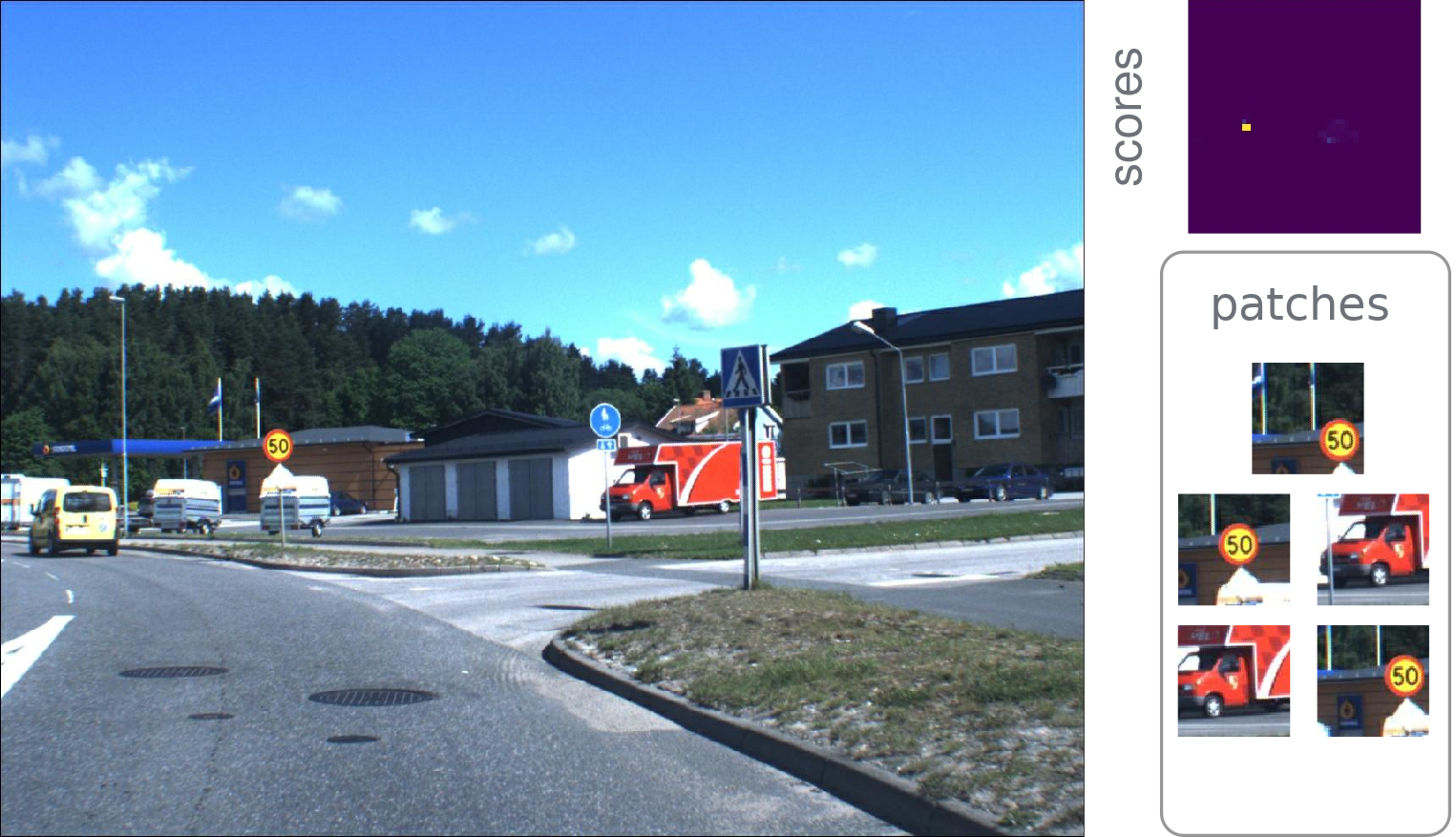}%
    \caption{\emph{Left:} Example of a 960$\times$1280 image from the traffic signs dataset. \emph{Top-right:} scores predicted for patch extraction. \emph{Right:} Patches extracted by perturbed Top-K at inference.}
    \label{fig:trafficsigns}
\end{figure}

\subsection{Inter-patch reasoning}

\begin{figure}
    \centering
    \includegraphics[width=\linewidth]{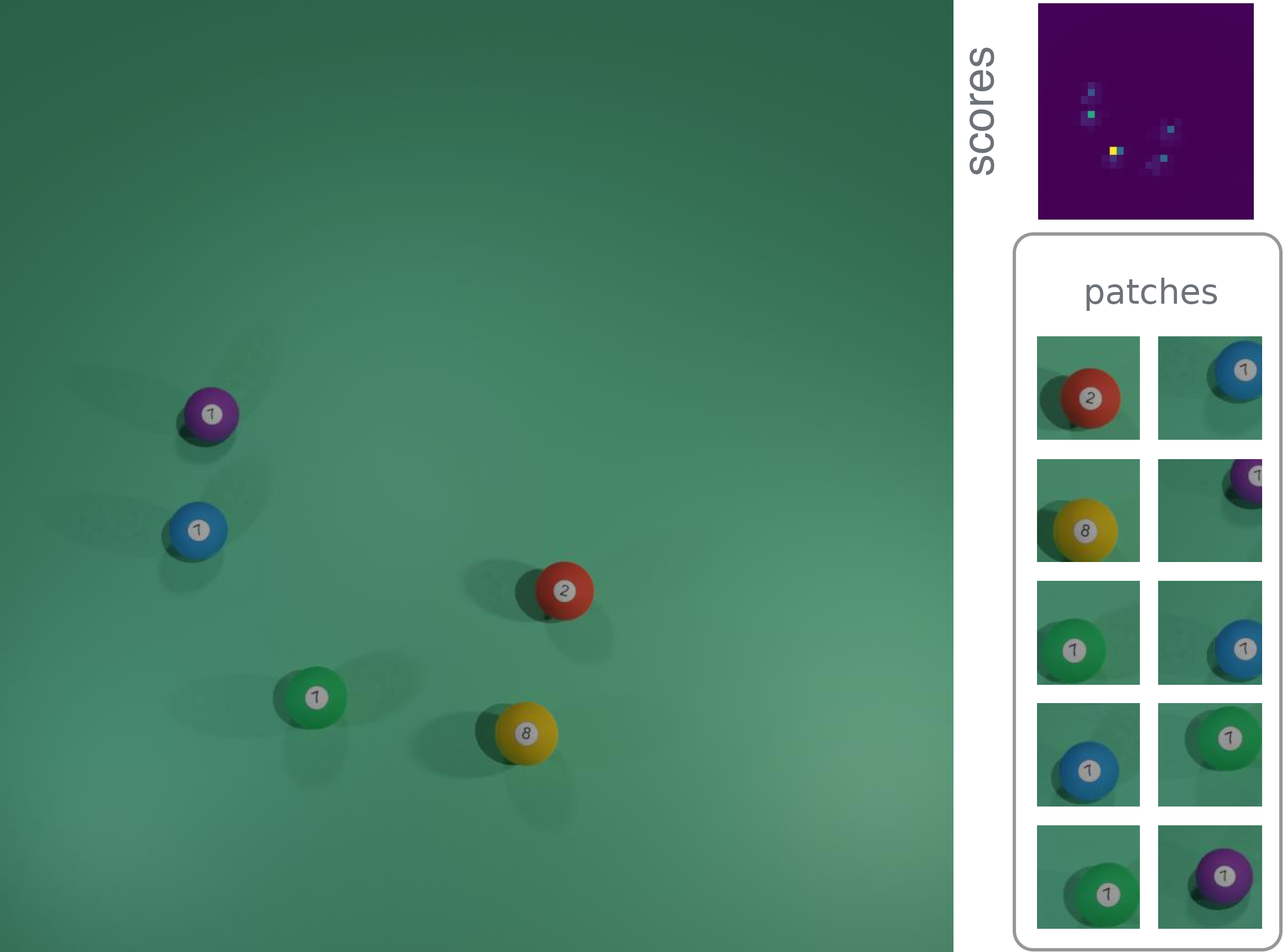}%
    \caption{\emph{Left:} example image from the billiard balls dataset. \emph{Top-right:} scorer network output for patch selection. \emph{Right:} Patches extracted at inference.}
    \label{fig:billiard}
\end{figure}

The work most closely related to ours, ATS, requires averaging representations obtained from sampled patches.
We hypothesize that such mean aggregation limits expressivity required for modelling relationships between extracted patches.
We investigate this using a synthetic dataset inspired by MegaMNIST~\cite{attention_sampling}, but which goes beyond scattered MNIST numbers on a Megapixel image to instead consist of billiard balls as presented in~\Cref{fig:billiard}. Each image contains four to eight randomly colored balls randomly placed on the table. Ball numbers are sampled uniformly from $\{1,\dots, 9\}$ and face the camera. We ensure that balls do not completely obstruction each other.
We define the following classification task: report the higher of two numbers extracted from the leftmost and rightmost balls. All the other balls can be ignored.
This task exhibits three interesting properties: (i) the information on the image is very localized (around the balls), (ii) downsampling the image severely degrades the readability of ball numbers, (iii) the task may be difficult to solve using a simple mean-pooling approach.

Using the Kubric\footnote{\url{https://github.com/google-research/kubric}} software, we generated 20k images of size 1000$\times$1000. We split them into 8k samples for training, 2k for validation, and 10k for test. The generated dataset is available for download\footnote{\url{http://storage.googleapis.com/gresearch/ptokp_patch_selection/billiard.tar.xz}}. Code for our experiments is available on GitHub\footnote{\url{https://github.com/google-research/google-research/tree/master/ptopk_patch_selection}}.

We apply our model to this task. Architecture details are as follows. The scorer is a 4-layer CNN and processes the downscaled image of size $250\times 250$ where the balls are visible but the numbers cannot be read.
The feature network is ResNet18~\cite{he2016deep}. We compare different aggregation schemes: mean-pooling, max-pooling, and a small Transformer~\cite{att_is_all_you_need}. The latter consists of 3 self-attention layers with 8 heads, taking the sequence of patch representations as input augmented with a learned additive positional encoding.
We compare against ATS as well as a simple ResNet18 baseline.
All the methods are trained using Adam~\cite{kingma2017adam} with decoupled weight decay of $10^{-4}$~\cite{loshchilov2019iclr} and tuned learning rate $10^{-4}$.
We repeated each experiment several times, but noted that not all runs were successful. For ATS, only 1 out of 4 runs were able to meaningfully solve the problem by reaching an accuracy of 53.25\,\%, while the remaining three attempts performed no better than predicting the majority class. We also tried concatenating $[0, 1]$ normalized fixed positional encodings as two extra channels in the input (`concat. position') as the task requires the model to consider ball position. This affords a fairer comparison between our Transformer variant and the ATS and ResNet baselines. Positional encoding considerably improved baseline performances but were not as effective as the transformer. With concatenated position encodings, 3 out of 4 ATS runs meaningfully solved the problem while 1 run always predicted the majority class. This failure mode of majority class prediction also happened to 2 out of 9 runs of our model when using max-pooling aggregation. Both mean-pooling and transformer aggregation were relatively stable. We report robust estimates of performance (median and median absolute deviation) in~\Cref{tab:billiard_results}.

\begin{figure}
    \centering
    \includegraphics[width=0.8\linewidth]{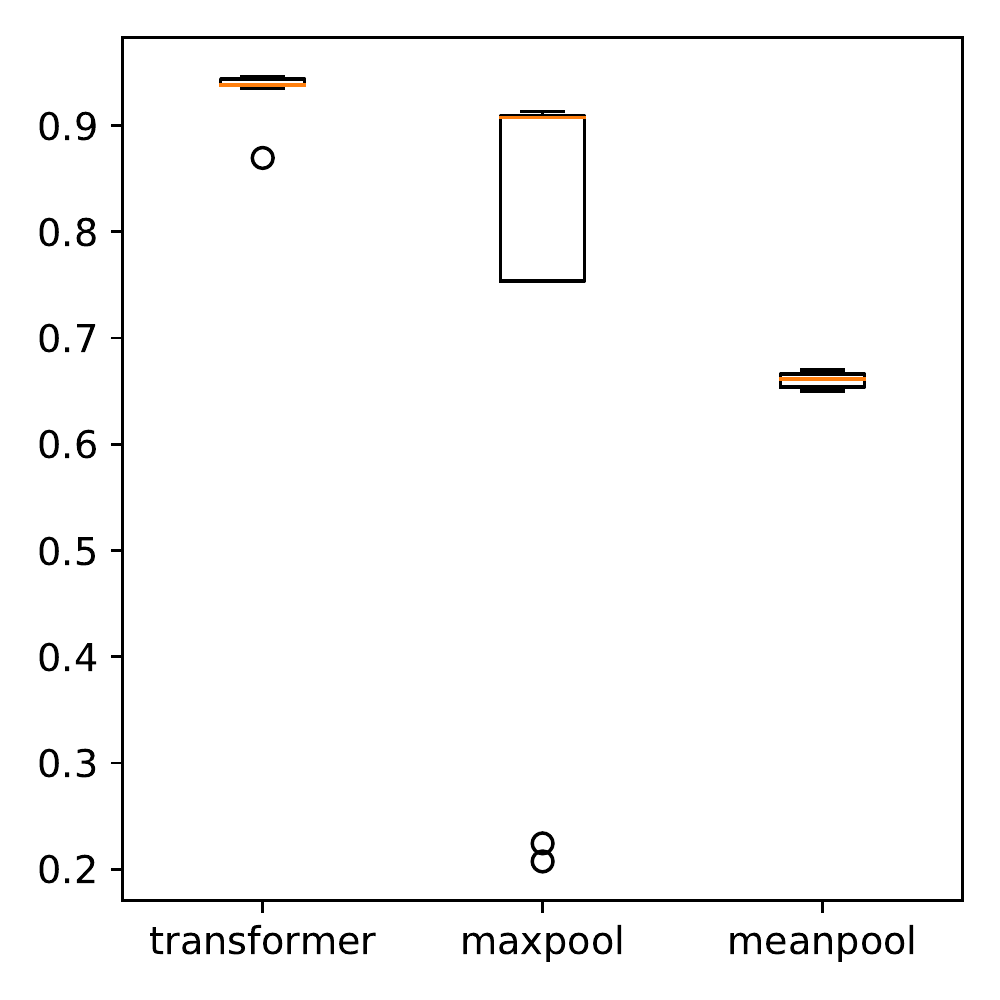}
    \caption{The effect of aggregation method in our model is ablated using 9 repetitions for each setting. Transformer and mean aggregation are relatively stable and transformer performs best.}
    \label{fig:cvpr21:cr:aggregationablation}
\end{figure}

As the results show, a standard CNN is able to solve this task, but the training time increase by 24\,\% compared to differentiable Top-K with a transformer on top. This difference in running time would be even more pronounced in higher resolutions (see supplementary material). One may reduce CNN training time by downsampling the input image. As demonstrated in the results, this does not work, as the numbers become too small to be readable. ATS is not able to solve this task reliably, most likely because of mean pooling per-patch embeddings. Our own method also performs poorly when using mean-pooling, but when using transformers and max-pooling aggregation we are able to solve this task with high accuracy, outperforming all other methods. We compare the three aggregation schemes using a box-whiskers plot in~\cref{fig:cvpr21:cr:aggregationablation}.

\begin{table}
\centering
\begin{tabular}{@{}lcc@{}}
\toprule
              &  pooling          & Test Acc. [\%]\\
\midrule
 & transformer & $93.8 \pm 0.3$\\
Top $K=10$ & max         & $90.8 \pm 0.1$ \\
  & mean       & $66.1 \pm 0.7$ \\
\midrule
ATS-10 \cite{attention_sampling} & mean & $20.7\pm 0.00$\\
\multicolumn{2}{l}{~~~~~ + concat. position}& $42.44\pm 10.85$\\
\midrule
ResNet18 & & $81.6 \pm 0.35$\\
\multicolumn{2}{l}{~~~~~ + concat. position}& $83.8 \pm 0.30$\\
\multicolumn{2}{l}{~~~~~ + downscale by 4} & $17.1 \pm 0.81$ \\
Majority class & & 21.0\phantom{$\pm 0.00$}\\
\bottomrule
\end{tabular}
\caption{Predicting the highest number between the rightmost and leftmost ball on the billiard balls dataset. Median and Median Absolute Deviation (MAD) for five runs (except for ATS where we have 4 runs and our method where we have 9 runs) of the Top-1 accuracy is reported.}
\label{tab:billiard_results}
\end{table}

\subsection{Fine-grained bird classification}
\label{ssec:birds}

\begin{table}[t]
\centering
\begin{tabular}{@{}lc@{}}
\toprule
Method & acc [\%] \\
\midrule
ResNet50 & 84.5\\
\midrule
MG-CNN \cite{Wang2015_ICCV} & 81.7\\
Bilinear-CNN \cite{Lin2015_ICCV} & 84.1\\
Spatial Transformer \cite{spatial_transformer} & 84.1\\
FCAN \cite{DBLP:journals/corr/LiuXWL16} & 84.5\\
PDFR \cite{Zhang_CVPR_PDFR} & 84.5\\
RA-CNN \cite{Fu2017_CVPR} & 85.3\\
HIHCA \cite{Cai2017_ICCV} & 85.3\\
Boost-CNN \cite{Moghimi2016} & 85.6\\
DT-RAM \cite{Li2017_ICCVW} & 86.0 \\
MA-CNN \cite{Zheng_2017_ICCV} & 86.5\\
\midrule
NTS-Net $k=2$ \cite{learning_navigate} & 87.3\\
NTS-Net $k=4$ \cite{learning_navigate} & 87.5\\
\midrule
NTS-Top-K $k=2$ (ours) [Concat] & $84.7 \pm 0.8$\\
NTS-Top-K $k=4$ (ours) [Concat] & $85.9 \pm 0.4$\\
NTS-Top-K $k=2$ (ours) [Mean] & $85.8 \pm 0.3$\\
NTS-Top-K $k=4$ (ours) [Mean] & $86.7 \pm 0.4$\\
\bottomrule
\end{tabular}
\caption{Results on \birds{} dataset. Error bars are standard deviation over 5 repeats. `Concat' and `mean' refer to the aggregation method. NTS-Net uses `Concat'. Numbers for prior work were taken from Table 2 of~\cite{learning_navigate}.}
\label{tab:birds_results}
\end{table}

Another way to extract salient regions of the image is to use a teacher-student approach to learn how to rank patches~\cite{learning_navigate} (NTS-Net).  We argue that NTS-Net's ranking loss is essential only due to the non-differentiability of their patch selection method, and show here how their model can be simplified using our differentiable Top-K module. We explore this using the Caltech-UCSD Birds (CUB-200) dataset~\cite{Wah2011_CUB200}, a common dataset for fine-grained image classification. This dataset is a bird classification task with 11,788 images from 200 bird species.

We briefly describe the training setup of NTS-Net: The original image is resized and cropped to  $(448 \times 448)$, from which ResNet activations $R_I$ are computed. A scorer network processes these activations to score patches at multiple scales and aspect ratios. The scorer is trained using a ranking loss which we skip here because it is not required in our formulation. $K$ top scoring (Top-K) patches are selected post greedy non-maxima suppression, resized to $(224 \times 224)$, and encoded using the same ResNet backbone. Call these activations $R_{P_1}, \cdots, R_{P_K}$. NTS-Net tries to predict the class label from $R_I$, each of $R_{P_j}$, and a concatenation of all these embeddings $\left[R_I, R_{P_1}, \cdots, R_{P_K}\right]$. The model minimize a softmax cross entropy loss on each prediction head.

In this section, the emphasis is not on efficiently processing high resolution input but rather on being able to process salient parts of the object at a level of detail that is computationally infeasible when processing the entire input.
We adapt out architecture to closely match NTS-Net, modulo four differences. First, NTS-Net uses greedy non-maximum suppression (NMS) in its patch selection module. Hard greedy NMS cannot be made differentiable using the perturbed maximum method as it does not correspond to the optimum of any linear program. Non-greedy-NMS, on the other hand, can be easily incorporated into the constraint set (\cref{eqn:constraitset}). It would be computationally infeasible to solve the resulting linear program $n=500$ times in each forward pass. One could learn a network to do NMS~\cite{learning2017cvpr} to overcome this problem. We instead resort to making our scorer network more expressive using Squeeze-Excitation layers~\cite{hu2018cvpr}, so that it may learn to select non-overlapping patches if necessary. This way of doing feature modulation enables global communication between all spatial locations, allowing us to model a crude form of global context.
Second, we use entropy regularization with a coefficient of $0.05$ as in other experiments above~\footnote{Regularization is applied on the softmax of concatenated unnormalized scores. The final RPN outputs are normalized to lie between $[0, 1]$ as in the other experiments.}. Third, we do not need and therefore eschew the ranking loss. Instead gradients can flow from the per-patch ResNet into the region proposal head through our differentiable patch selection module. Lastly, two aggregation methods are explored: (a) `concat' matching that in NTS-Net and (b) `mean' $R_I + \frac{\left(R_{P_1} + \cdots + R_{P_K}\right)}{K}$ where per-patch embeddings are averaged and added to the whole image embedding. `Concat' results in a massive linear classifier over excessively high dimensional embeddings. `Mean' ensures that aggregated embeddings are $2048$ dimensional.

We tuned optimization hyper-parameters and regularization strength by splitting the training set into 5000 training images and 994 validation images. This was to avoid meta-overfitting on test data. We took the best hyper-parameter combination from this train-val split and then trained the model on the entire 5994 training images and tested on the official test set. Test results are reported in Table~\ref{tab:birds_results}.

We note that our method performs slightly worse than the NTS-Net baseline. This may be due to the following: (a) Our model may be more prone to overfitting. It can select patches that directly optimize the training loss. We combat this using `mean' aggregation which improves performance as shown in the last two rows in Table~\ref{tab:birds_results}. (b) We report average performance across 5 seeds. Our best performing `mean'-aggregation model achieves 87.3\%. (c) Despite using SE modules in the scorer network, our model still selects overlapping patches around the bird head. This likely also makes overfitting worse. Please see supplementary material for qualitative results.

\section{Discussion and future work}
We introduced a differentiable Top-K module for patch selection in large images.
This enabled end-to-end learning of the patch selection module in a task dependent manner.
We observed competitive performance on a task where information relevant for classification was very localized within the image.
We significantly advanced the state-of-the-art in a problem requiring inter-patch relationship reasoning, as evidence by our use of a Transformer module on patch embeddings in the billiard balls dataset. Our method allows for arbitrary downstream neural processing of patches without using auxiliary losses for training the scorer network.

We are delighted by the above progress but note that there is still significant room for improvement and identify the following areas for future work. The patch selection problem is a chicken-and-egg problem. One can pick optimal patches by first knowing the contents of the image, but to efficiently know the contents of the image one should pick the right patches. This makes it inherently difficult to properly account for global context. Training patch selection models remains difficult. Some failure modes include predicting the majority class, or the scorer becoming very confident of the wrong patches either early in training or diverging to this behavior in the middle of training. More adaptive optimization might help improve learning stability. Patch selection lead to less overfitting on the Traffic street signs dataset. However, the same model overfits on the~\birds~dataset. The scorer was able to identify patches associated with watermarks and background in order to memorize the training set. Fortunately when combined with the NTS-Net framework, overfitting was considerably reduced. Future work could address this by training on larger dataset or using self-supervised learning.

\section*{Acknowledgements}
We thank Thomas Kipf, Francesco Locatello, Georg Heigold, Klaus Greff, Sindy Löwe, and Quentin Berthet for helpful discussions.

{\small
\bibliographystyle{ieee_fullname}
\bibliography{references}
}

\clearpage

\appendix

\section{Supplemental Material}

The supplementary material consists of the following: performance trade-offs associated with patch sampling versus running a CNN on the entire high resolution image (\Cref{sec:overhead}), some theoretical discussions regarding the LP formulation of index-sorted Top-K in (\Cref{appx:lp}), an experiment quantifying the effect of decaying $\sigma$ to 0 during training in (\Cref{appx:decays0}), results with another differentiable top-K method we experimented with before developing our approach (\Cref{sec:differentiable_sinkhorn}), hyper-parameter details for all our experiments (\Cref{appx:hparams}), qualitative results (\Cref{appx:qualresults}), and a PyTorch implementation of the perturbed Top-K module (\Cref{appx:pytorchcode}).

\section{Speed Improvements by Sampling Patches}\label{sec:overhead}
\begin{figure}[t!]
    \centering
    \includegraphics[width=\linewidth]{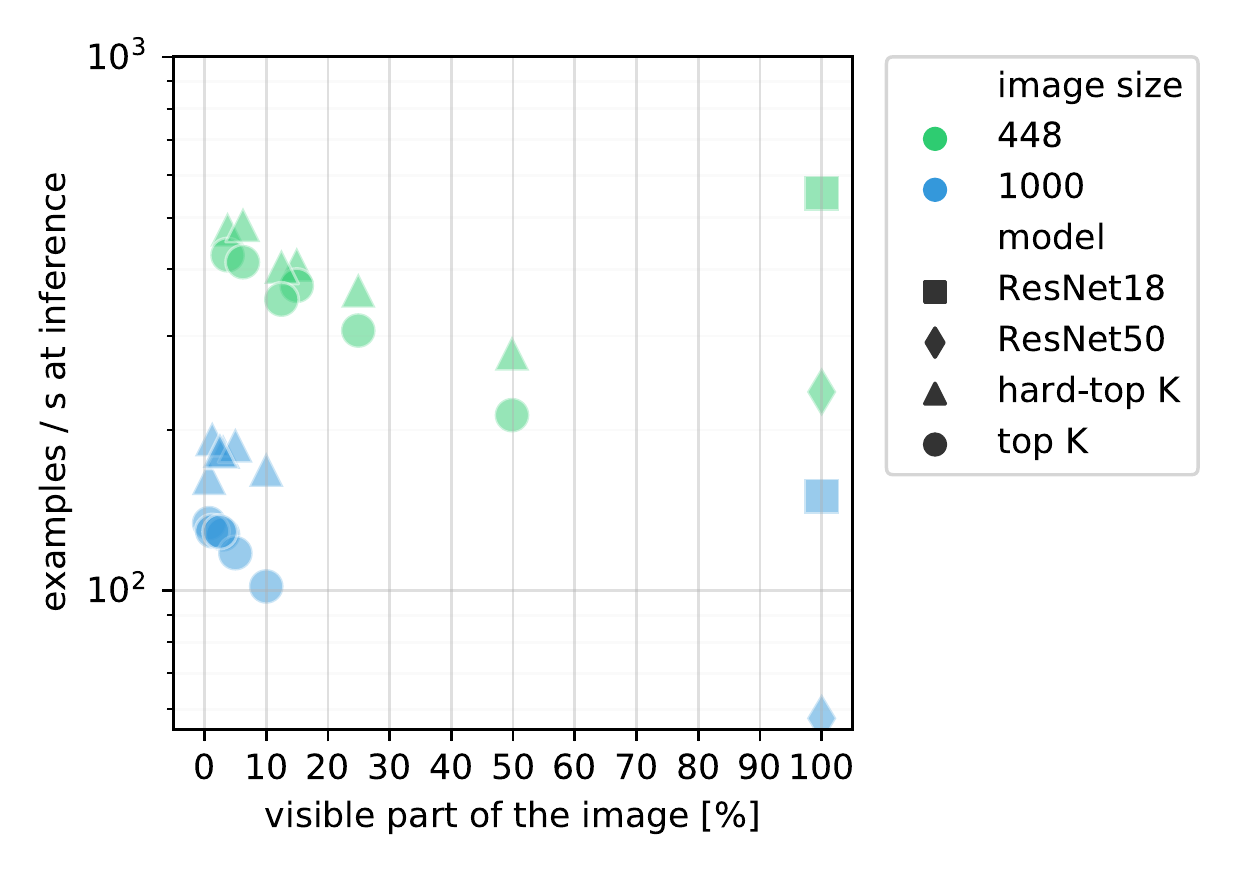}
    \caption{Comparison of inference speed on a single V100 between ResNet 50 at full resolution images vs. extracting only $K \in \{3, 5, 10\}$ patches of dimension $P \in \{50, 100\}$ with differentiable or hard Top-K. The feature network is a ResNet50 and aggregation is mean-pooling. $x$-axis is the percentage of extracted pixels $\left(K\cdot P^2\right)$ compared to the full image. For each point in this plot, the maximum over feasible batch-sizes was used to obtain the highest possible throughput. }
    \label{fig:patch_tradeoff}
\end{figure}

We study the speed improvement that can be gained at inference by using our patch extraction model compared to running a model on the full image.
We compare the number of samples processed per second at inference on a single V100 GPU in Figure~\ref{fig:patch_tradeoff}.
If the useful information for recognition is localized within than 10\% of the pixels, which corresponds to extracting $K=10$ patches of size $100\times 100$ on a MegaPixel image, processing only the relevant regions with the same network (ResNet50) allows a 5 fold speed up. In this case, roughly 28\,\% of the inference time is spent on the feature network, while most of the remaining time is spent calculating the scores for the individual patches. The Top-K operation's influence on runtime is minimal.
This should be put in contrast with the common alternative of running a significantly smaller ResNet18 on the full size image which is not faster on $1000\times 1000$ images than our approach running a ResNet50 on the extracted patches.
Using hard Top-K at inference instead of the differentiable version is consistently faster on a V100 GPU.

\section{Linear Program}\label{appx:lp}
To understand why the LP, in equations 4 and 5, corresponds to index sorted top-K, let us focus on the integral solutions $\mY_{ik} \in \{0, 1\}$. In this scenario, each column of $\mY$ is a one-hot indicator vector selecting exactly one of the scores in $s$. Furthermore, the objective function is the sum of selected scores. Thus the optimal solution is to select the $K$ highest scores.

Lastly, the strict inequality constraint forces the selection of numbers with lower indices earlier. This can be proven by contradiction: \\
Consider a valid solution which has $l > m$, but $s_l$ was selected by the $k^{\mathrm{th}}$ column and $s_m$ by $k'^{\mathrm{th}}$ column such that $k < k'$. That is, $\mY[:, k] = \mathrm{one\,hot}(l)$ and $\mY[:, k'] = \mathrm{one\,hot}(m)$. Then, since this solution is valid the following holds
\begin{align}
\sum_{i\in[N]} i\mY_{i,k} <& \sum_{j\in[N]} j\mY_{j,k'} \\
\implies l <& m
\end{align}
which is a contradiction to the original assumption.

\section{Decaying $\sigma$ to $0$}\label{appx:decays0}
\begin{figure}
    \centering
    \includegraphics[width=0.8\linewidth]{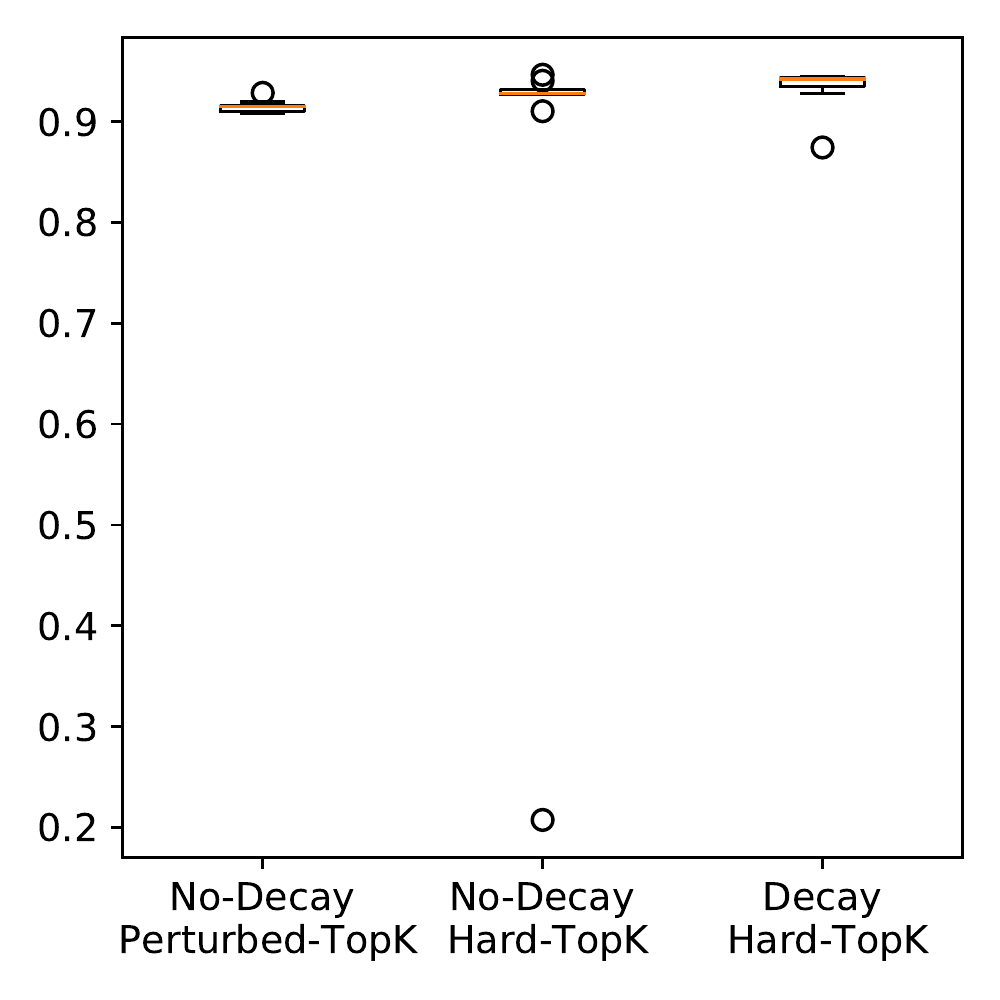}
    \caption{Ablating the effect of reducing $\sigma$ (perturbation magnitude) to 0 during training. Each setting is repeated 9 times. \textit{Left:} $\sigma$ constant and perturbed top-K for both training and inference. \textit{Middle:} $\sigma$ constant, perturbed top-K for training and hard top-K for inference. \textit{Right:} $\sigma$ decays to 0 during training, perturbed top-K during training and hard top-K for inference.}
    \label{fig:cvpr21:cr:billiarddecayabl}
\end{figure}

We linearly decay perturbation magnitude $\sigma$ to $0$ in all our models. We ablate this design choice on the billiard balls dataset. These experiments use the transformer aggregation head. All other hyper-parameters are kept the same. In~\cref{fig:cvpr21:cr:billiarddecayabl} and~\cref{tab:sigma0ablation} we see that using hard top-k at inference leads to higher accuracy ($+1.3\%$). Decaying perturbation magnitude to zero further improves performance ($+1.4\%$).

\begin{table}[h]
\centering
\begin{tabular}{@{}lc@{}}
\toprule
                          & test acc. [\%]\\
\midrule
No-Decay, Perturbed-TopK  & $91.5 \pm 0.4$\\
No-Decay, Hard-TopK       & $92.8 \pm 0.4$\\
Decay, Hard-TopK          & $94.2 \pm 0.2$\\
\bottomrule
\end{tabular}
\caption{Ablating the effect of reducing $\sigma$ (perturbation magnitude) to 0 during training. See caption in~\cref{fig:cvpr21:cr:billiarddecayabl} for details.}
\label{tab:sigma0ablation}
\end{table}

\section{Differentiable Sinkhorn}\label{sec:differentiable_sinkhorn}

Another approach to make Top-K differentiable was proposed by Xie~\emph{et. al}~\cite{smooth_topk_xie} and relies on the optimal transport formulation of Top-K proposed by Cuturi~\emph{et. al}~\cite{diff_rank_sort_cuturi}. We implemented the forward and backward pass following Algorithm 3 of~\cite{smooth_topk_xie}.
We report the results for traffic sign recognition in Table~\ref{tab:trafficsigns_results_sinkhorn} with similar setting as in Section 5.1. This approach gives good results but suffers when using discrete Top-K at inference.

\begin{table}[h]
\centering
\begin{tabular}{@{}lc@{}}
\toprule
                        & test acc. [\%]\\
\midrule
Sinkhorn Top $K=5$                 & $95.4 \pm \phantom{1}0.7$ \\
~~~+ discrete Top-K at inference   & $83.0 \pm 10.5$ \\
Sinkhorn Top $K=10$                & $92.6 \pm \phantom{1}3.1$ \\
~~~+ discrete Top-K at inference   & $86.3 \pm \phantom{1}1.7$ \\
\bottomrule
\end{tabular}
\caption{Performance of Sinkhorn Top-K based models on the traffic signs dataset. We report the mean and standard deviation across 5 runs.}
\label{tab:trafficsigns_results_sinkhorn}
\end{table}

\begin{table*}[ht!]
\centering
\begin{tabular}{llll}
\toprule
 Dataset & Traffic Signs & Billiard balls & CUB-200\\
\midrule
 Batch Size & 32 & 64 & 16 \\
 Learning rate (LR) & $10^{-4}$ & $10^{-4}$ & $10^{-3}$ \\
 Weight decay & $10^{-4}$ & $10^{-4}$ & $10^{-4}$ \\
 Steps & 100\,000 & 30\,000 & 31\,300 \\
 LR Schedule & Cosine decay & \multirow{2}{1.5in}{Cosine decay + 5\% warm-up} & \multirow{2}{2in}{Piece-wise constant + Decay by 0.1 at 60 and 80 epochs + 5\% warm-up} \\
 & & & \\
 Optimizer & Adam-W & Adam-W & SGD + Momentum \\
 Entropy regularizer & 0.01 & 0.01 & 0.05 \\
 Gradient clip value & 0.1 & 1.0 & - \\
\bottomrule
\end{tabular}
\caption{Summary of optimization hyper-parameter settings.}
\label{tab:optimizer}
\end{table*}

\begin{figure*}[h!]
     \centering
     \begin{subfigure}[b]{0.4\textwidth}
         \centering
         \includegraphics[width=\textwidth]{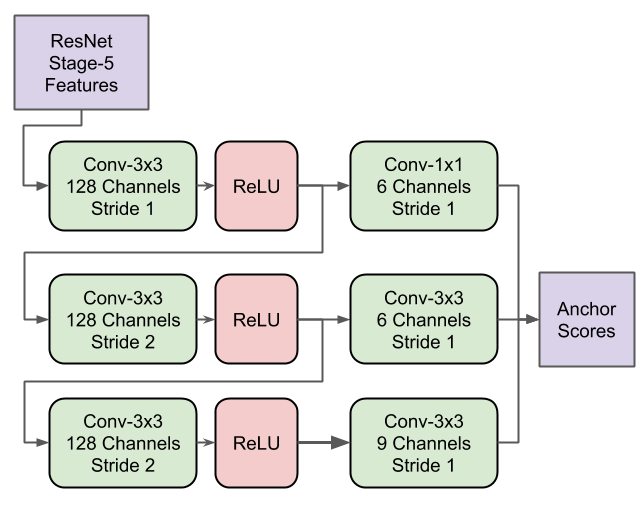}
         \caption{NTS-Net}
         \label{fig:ntsnet}
     \end{subfigure}
     \begin{subfigure}[b]{0.59\textwidth}
         \centering
         \includegraphics[width=\textwidth]{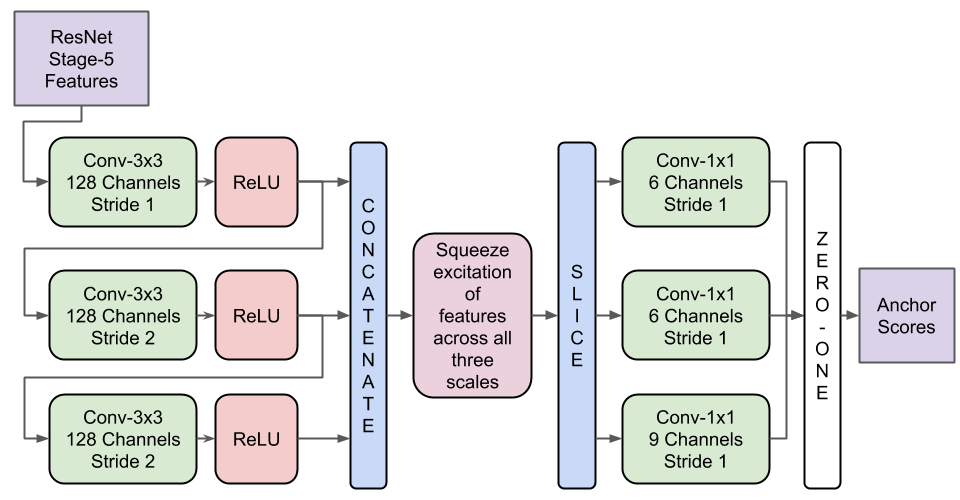}
         \caption{Top-K NTS-Net}
         \label{fig:ptopkntsnet}
     \end{subfigure}
     \caption{Architecture of the region proposal head used in the baseline NTS-Net model (a), and in our adaptation (b), for fine-grained classification in the \birds~dataset. We added a squeeze excitation layer in the middle to allow for communication between all features just before anchor prediction. Zero-one normalization is applied across all 1614 scores.}
     \label{fig:birdsrpn}
\end{figure*}

\section{Experimentation Details}\label{appx:hparams}
For all experiments except the Fine-grained bird classification ones, our scorer network consisted of a CNN with 4 convolutional layers of kernel size $3 \times 3$ with stride 1 and ``valid'' padding. The number of feature maps was 8, 16, 32 and 1, respectively. Every convolution except for the last was followed by a Relu activation function. The last convolution was followed by an $8 \times 8$ max pooling of stride 8.

Our feature network was different depending on the dataset: On the Swedish Traffic Signs dataset, we used the same feature network as ATS~\cite{attention_sampling}, which is a narrow ResNet18 with 16 filters instead of the usual 64, 128, 256, 512. On the billiard balls dataset we used a standard ResNet18. On the \birds~dataset we use the same feature extractor as NTS-Net, that is a ResNet50.

We used the ADAM-W optimizer~\cite{loshchilov2019iclr} for Traffic signs and billiard balls dataset. We used SGD with momentum $0.9$, similar to NTS-Net, for \birds. The exact details of our optimizers can be seen in~\Cref{tab:optimizer}. We found that weight decay coupled with momentum updates was better to reduce overfitting on \birds.

In our adaptation of the NTS-Net, we added Squeeze Excitation layers inside the region proposal network (RPN). This was meant to compensate for the lack of non-maximum suppression in our patch selection module. The RPN for both our method and the baseline NTS-Net are shown in figure~\ref{fig:birdsrpn}. Lastly, with regards to data pre-processing, instead of normalizing pixels by a pre-computed pixel mean and pixel standard deviation, we re-scaled pixel values to lie between $[-1, 1]$ during training. This was to match the value range we used for pre-training our ResNet50 backbone on ImageNet ILSVRC12. Other than this minor change, we used the same data augmentation as NTS-Net.

\section{Qualitative Results}\label{appx:qualresults}
\begin{figure*}[t!]
    \centering
    \includegraphics[width=\linewidth]{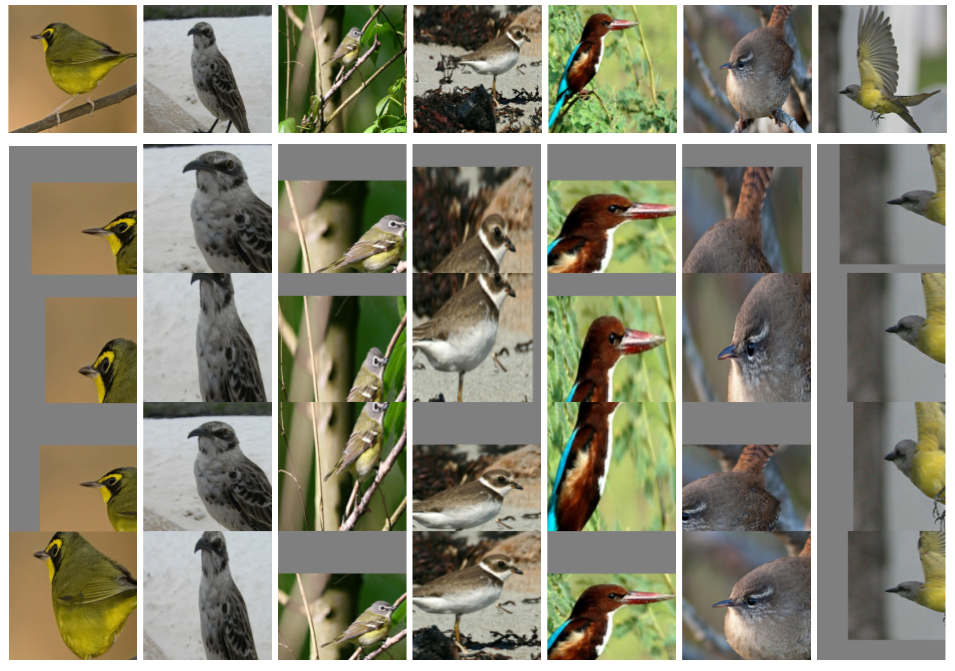}
    \caption{Patches extracted from images in the test split of the \birds~dataset. The first row shows the original input image after being resized to $600\times600$ followed by a centre crop of $480\times480$. Padding artifacts correspond to selecting anchors that include pixels outside the image boundary.}
    \label{fig:birdspatches}
\end{figure*}

We visualize patches extracted by the model on the \birds~dataset, on the test split, in~\cref{fig:birdspatches}. This particular model is the best performing seed among the 5 random repeats we averaged to report the number for $K=4$ in the main manuscript. It uses `mean' aggregation and achieves $87.3\%$ top-1 accuracy. The images visualized are not cherry picked. We see that the model is able to focus on the bird and captures the region around the eyes and torso. The lack of NMS is evident as the model often looks at patches with high overlap.

\clearpage
\onecolumn
\section{Differential Top-K PyTorch Code}\label{appx:pytorchcode}
\begin{lstlisting}[language=Python]
    import torch
    import torch.nn as nn

    class PerturbedTopK(nn.Module):
        def __init__(self, k: int, num_samples: int = 1000, sigma: float = 0.05):
            super(PerturbedTopK, self).__init__()
            self.num_samples = num_samples
            self.sigma = sigma
            self.k = k
    
        def __call__(self, x):
            return PerturbedTopKFunction.apply(x, self.k, self.num_samples, self.sigma)

    class PerturbedTopKFunction(torch.autograd.Function):
        @staticmethod
        def forward(ctx, x, k: int, num_samples: int = 1000, sigma: float = 0.05):
            b, d = x.shape
            # for Gaussian: noise and gradient are the same.
            noise = torch.normal(mean=0.0, std=1.0, size=(b, num_samples, d)).to(x.device)
    
            perturbed_x = x[:, None, :] + noise * sigma  # b, nS, d
            topk_results = torch.topk(perturbed_x, k=k, dim=-1, sorted=False)
            indices = topk_results.indices  # b, nS, k
            indices = torch.sort(indices, dim=-1).values  # b, nS, k
    
            # b, nS, k, d
            perturbed_output = torch.nn.functional.one_hot(indices, num_classes=d).float()
            indicators = perturbed_output.mean(dim=1)  # b, k, d
    
            # constants for backward
            ctx.k = k
            ctx.num_samples = num_samples
            ctx.sigma = sigma
    
            # tensors for backward
            ctx.perturbed_output = perturbed_output
            ctx.noise = noise
    
            return indicators
    
        @staticmethod
        def backward(ctx, grad_output):
            if grad_output is None:
                return tuple([None] * 5)
    
            noise_gradient = ctx.noise
            expected_gradient = (
                torch.einsum("bnkd,bnd->bkd", ctx.perturbed_output, noise_gradient)
                / ctx.num_samples
                / ctx.sigma
            )
            grad_input = torch.einsum("bkd,bkd->bd", grad_output, expected_gradient)
            return (grad_input,) + tuple([None] * 5)
\end{lstlisting}

\end{document}

%% file: math_commands.tex
\usepackage{amsmath,amsfonts,bm}

\def\eqref#1{equation~\ref{#1}}

\def\1{\bm{1}}

\def\vone{{\bm{1}}}

\def\vs{{\bm{s}}}

\def\vy{{\bm{y}}}

\def\mH{{\bm{H}}}

\def\mS{{\bm{S}}}

\def\mY{{\bm{Y}}}

\DeclareMathAlphabet{\mathsfit}{\encodingdefault}{\sfdefault}{m}{sl}
\SetMathAlphabet{\mathsfit}{bold}{\encodingdefault}{\sfdefault}{bx}{n}
\newcommand{\tens}[1]{\bm{\mathsfit{#1}}}

\def\tP{{\tens{P}}}

\def\tX{{\tens{X}}}

\newcommand{\E}{\mathbb{E}}

\newcommand{\R}{\mathbb{R}}

\DeclareMathOperator*{\argmax}{arg\,max}